\def\paperlanguage{}
\pdfoutput=1 

\documentclass[letterpaper, 10 pt, conference]{ieeeconf}  

\usepackage{bm}
\usepackage{cite}
\usepackage{flushend}
\include{preamble}

\IEEEoverridecommandlockouts                              

\title{\LARGE \textbf
  {
    \switchlanguage%
    {%
      Adaptive Robotic Tool-Tip Control Learning Considering\\Online Changes in Grasping State
    }%
    {%
      逐次的な把持状態変化を考慮した適応的道具先端操作学習
    }%
  }
}

\author{Kento Kawaharazuka$^{1}$, Kei Okada$^{1}$, and Masayuki Inaba$^{1}$
  \thanks{$^{1}$ The authors are with the Department of Mechano-Informatics, Graduate School of Information Science and Technology, The University of Tokyo, 7-3-1 Hongo, Bunkyo-ku, Tokyo, 113-8656, Japan.
    {\texttt\small [kawaharazuka, k-okada, inaba]@jsk.t.u-tokyo.ac.jp}
  }
}

\begin{document}

\maketitle
\thispagestyle{empty}
\pagestyle{empty}

\begin{abstract}
  \switchlanguage%
  {%
    Various robotic tool manipulation methods have been developed so far.
    However, to our knowledge, none of them have taken into account the fact that the grasping state such as grasping position and tool angle can change at any time during the tool manipulation.
    In addition, there are few studies that can handle deformable tools.
    In this study, we develop a method for estimating the position of a tool-tip, controlling the tool-tip, and handling online adaptation to changes in the relationship between the body and the tool, using a neural network including parametric bias.
    We demonstrate the effectiveness of our method for online change in grasping state and for deformable tools, in experiments using two different types of robots: axis-driven robot PR2 and tendon-driven robot MusashiLarm.
  }%
  {%
    これまで様々な道具操作学習が行われてきた.
    しかし, そのどれもが道具操作中に逐次的に道具の把持位置・角度等が変化してしまうことを考慮していない.
    また 変形する道具を扱うことができる研究も少ない.
    本研究では, Parametric Biasを含むニューラルネットワークを利用し, 道具先端の位置推定・道具操作のための制御入力計算, 身体と道具の間の関係の変化への逐次的な適応を扱う手法を開発する.
    身体と道具の位置関係変化・変形する道具について本研究が有効であることを, 実機実験において示す.
  }%
\end{abstract}

\section{Introduction}\label{sec:introduction}
\switchlanguage%
{%
  Tool-use is one of the essential human abilities.
  So far, various studies have been conducted on the elements necessary for robotic tool-use, such as tool recognition \cite{huber2004template}, tool understanding \cite{zhu2015understanding}, tool selection \cite{tee2018tool}, tool manipulation \cite{toussaint2018tooluse}, and tool generation \cite{kawaharazuka2020tool}.
  Among these, robotic tool manipulation learning is one of the most important points in the actual robot operation.
  There are various stages of tool manipulation, including grasping a tool, understanding the positional relationship between the tool and the body, and planning the tool manipulation.
  For tool grasping, methods using kinematics and dynamics models \cite{miller2004graspit, xue2020gripping} and learning methods \cite{mahler2019dexnet} are considered.
  For the understanding of the positional relationship between the tool and the body, most of the methods \cite{hoffmann2014tooluse, nabeshima2006tool} obtain the linear transformation or Jacobian between the hand and tool postures.
  For tool manipulation planning, there are several methods such as optimization-based motion planning \cite{fang2018toolgrasp}, deep learning-based motion planning \cite{xie2019improvisation}, and methods using simple tool trajectory input and whole body inverse kinematics \cite{okada2006tool}.
  There are also methods to solve these problems simultaneously by imitation learning \cite{takahashi2017toolbody, saito2021tooluse}, reinforcement learning \cite{eppe2019tooluse}, and self-supervised learning \cite{fang2020tool, mar2018affordance}.

  However, none of them have taken into account the fact that the grasping state such as grasping position and tool angle can change at any time during the tool manipulation.
  In addition, there are few studies that can handle deformable tools.
  Previous studies have basically dealt with only the state in which a rigid tool is fixed to the robot body.
  Therefore, in this study, we develop a method that can handle both rigid and deformable tools by learning the relationship between the control command of the robot body and the tool-tip position using a neural network.
  By using parametric bias \cite{tani2002parametric, tani2004parametric}, the current grasping state, which cannot be obtained directly from the sensor information, is implicitly estimated online, and the control command for tool manipulation is changed based on the estimated grasping state.
  This system will be able to handle grasping states that can change at any time due to external forces, and deformable tools such as a long, flexible rod or a hose.
  We also apply our method to musculoskeletal humanoids \cite{wittmeier2013toward, kawaharazuka2019musashi}, which are flexible and more difficult to modelize for the grasping state.

  Note that the parametric bias \cite{tani2002parametric, tani2004parametric} is an additional bias parameter of neural network, which can extract multiple attractor dynamics from various motion data, mostly used in imitation learning.
  In the context of imitation learning, there are some examples to embed implicit tool differences into parametric bias \cite{nishide2012toolbody}.
  In this study, by using parametric bias instead of directly using the length and angle of the tool for a neural network, the need for annotation of the grasping state can be eliminated when creating the dataset, and deformable tools and complex grasping states can be handled.
  ``Grasping state'' in this study is defined as an implicit expression of various grasping states including grasping position, tool angle, etc., by parametric bias.
}%
{%
  道具操作は人間の本質的な能力の一つである.
  これまで, ロボットにおける道具使用に必要な要素である, 道具認識\cite{huber2004template}, 道具理解\cite{zhu2015understanding}, 道具選択\cite{tee2018tool}, 道具操作\cite{toussaint2018tooluse}, 道具生成\cite{kawaharazuka2020tool}に関して様々な研究が行われてきた.
  その中でも道具操作学習は実際のロボットにおける動作において最も重要である. 
  道具操作にも様々な段階があり, 道具の把持, 道具と身体の位置関係の理解, 道具操作のプランニングがある.
  道具の把持については, dynamics modelを使う方法\cite{miller2004graspit, xue2020gripping}や学習型の方法\cite{mahler2019dexnet}が考えられている.
  道具と身体の位置関係の理解では, 手先と道具の間の線形変換またはヤコビアンを求める場合\cite{hoffmann2014tooluse, nabeshima2006tool}がほとんどである.
  道具操作のプランニングには, optimization-based motion planning \cite{fang2018toolgrasp}や, deep learning-based motion planning \cite{xie2019improvisation}, simpleな道具軌道の入力とwhole body IKを使った方法\cite{okada2006tool}等がある.
  また, これらをひとまとまりにして模倣学習\cite{takahashi2017toolbody, saito2021tooluse}や強化学習\cite{eppe2019tooluse}, 自己教師あり学習\cite{fang2020tool, mar2018affordance}により解決する方法も存在する.

  しかし そのどれもが道具操作中に逐次的に道具の把持位置・角度等が変化してしまうことを考慮していない.
  また, 変形する道具を扱っている研究も少なく, 基本的には剛体の道具が身体に固定された状態のみを扱っている.
  そこで本研究では, ロボット身体の制御入力と道具の先端位置の間の関係をニューラルネットワークにより学習し, 変形する道具等も扱えるような手法を開発する.
  Parametric Bias \cite{tani2002parametric, tani2004parametric}を用いることでセンサ情報からは直接得られない現在の把持状態を逐次的に推定し, それを元に制御入力を変更していく.
  これらにより, 外力により逐次的に変化する把持状態, 重くてしなる棒や, ホースから出た水等も扱えるようになると考える.
  また, 筋骨格ヒューマノイド\cite{wittmeier2013toward, kawaharazuka2019musashi}等の柔軟でより把持状態のモデル化が難しいロボットにも適用を試みる.

  ここで, parameter biasとは, 主に模倣学習で使われる, 様々な動作データから複数のattractor dynamicsを抽出可能な, ニューラルネットワークにおける追加のbias parameterである.
  模倣学習の文脈では, 暗黙的な道具の違いを埋め込むために利用された例もある\cite{nishide2012toolbody}.
  本研究では, 直接道具の長さや角度を入力とせず, parametric biasを利用することで, データセット作成時の把持状態のアノテーションが必要なくなり, かつ変形する道具や複雑な把持状態を持つロボットを扱うことが可能となる.
  以降, 本研究では把持状態を, parametric biasを使った把持位置や把持姿勢等を含む様々な把持状態の暗黙的な表現として定義し用いる.
}%

\begin{figure}[t]
  \centering
  \includegraphics[width=1.0\columnwidth]{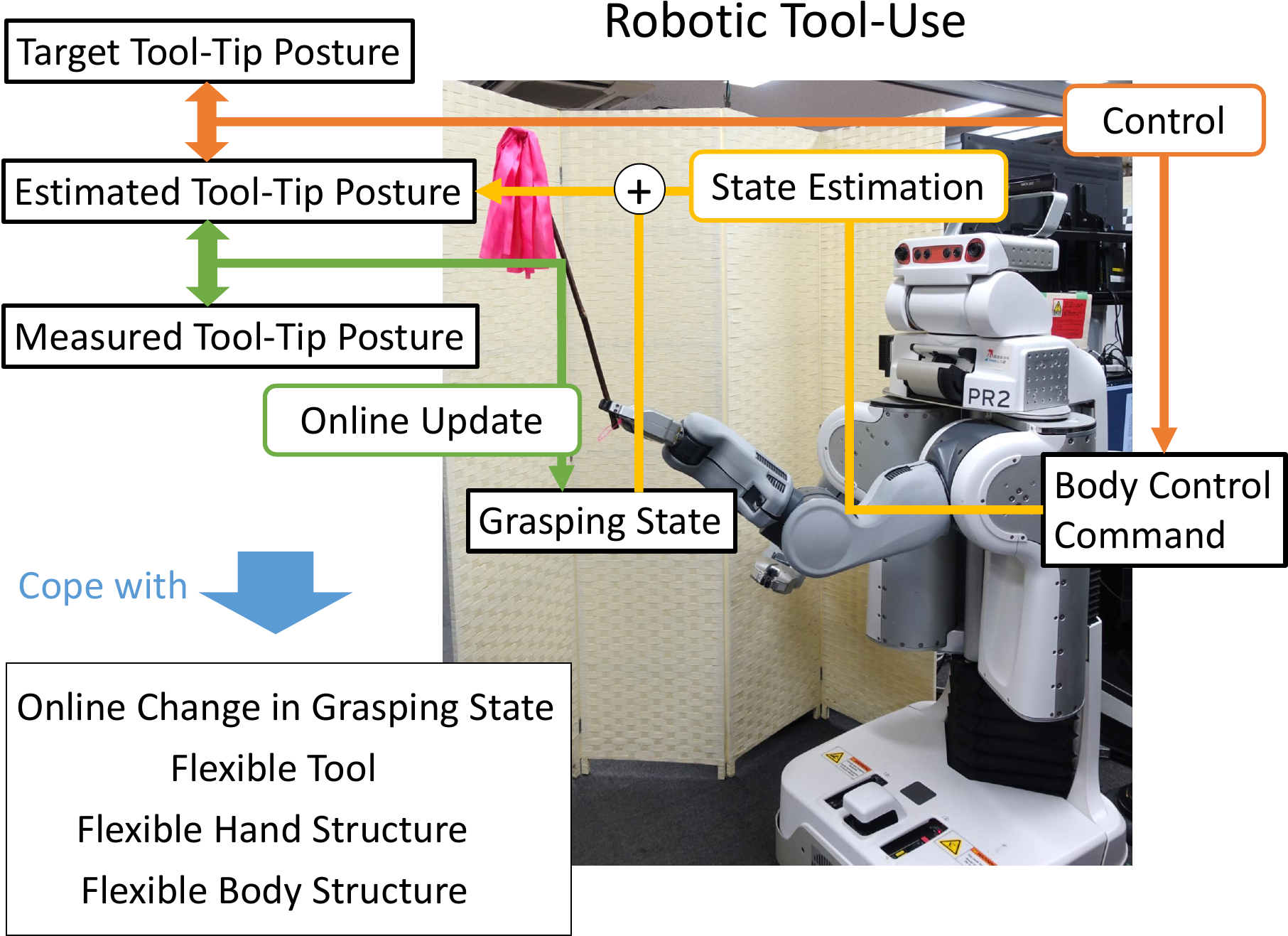}
  \vspace{-3.0ex}
  \caption{The concept of this study. In robotic tool-use, a tool-tip posture is estimated from the body control command and grasping state, the body control command is calculated from the loss between the target and estimated tool-tip postures, and grasping state is updated online from the loss between the estimated and measured tool-tip postures. This study can also cope with the online change in grasping state and flexible tool, hand, and body structures.}
  \label{figure:concept}
  \vspace{-3.0ex}
\end{figure}

\begin{figure*}[t]
  \centering
  \includegraphics[width=2.0\columnwidth]{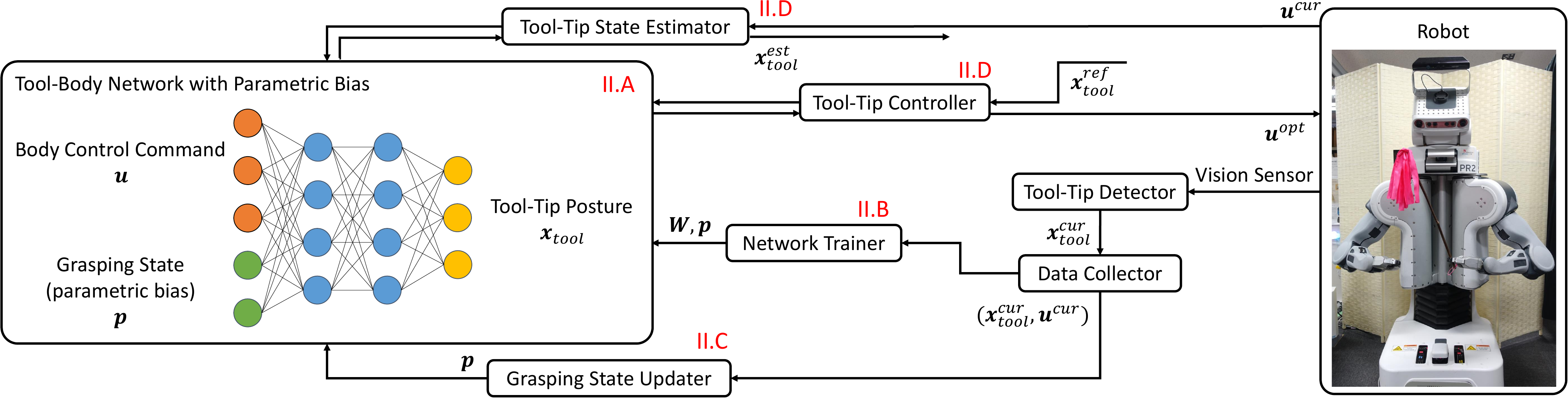}
  \vspace{-1.0ex}
  \caption{The overall software system: the network structure of TBNPB, network trainer of TBNPB, online grasping state updater through parametric bias, tool-tip state estimator, and tool-tip controller.}
  \label{figure:whole-system}
  \vspace{-3.0ex}
\end{figure*}

\switchlanguage%
{%
  Possible alternatives to our method are (1) a method using visual or tactile feedback, and (2) a method using a geometric model to estimate the grasping state.
  For (1), we can consider tactile feedback that can robustly respond to unexpected changes in the grasping state by storing or learning the sensor value transitions during tool-use \cite{pastor2012associative, girgin2018associative}, and visual feedback for the tool-tip position.
  For (2), a simple method to determine the grasping position and tool angle from the relationship between the hand position and the tool-tip position using a geometric model of the tool can be considered.
  We can say that (1) is a method to compensate the grasping state by sensor feedback without estimating it, and (2) is a method to use the tool by understanding the grasping state from the geometric model.
  However, (1) cannot deal with deformable tools and complex robot structures where Jacobian between the control command and the target state to be controlled is not obvious.
  In addition, the scope of application of (1) is different from that of this study because (1) mainly follows the human demonstration and does not modelize the tool or grasping state.
  There is also a tool-tip control with sensor feedback using imitation learning \cite{sasagawa2021autoregressive}, but there is no example of adaptation to changes in the grasping state of a tool.
  Since (2) assumes a geometric model, it cannot handle deformable tools or complex grasping states.
  In contrast, this study provides a general-purpose model that can be applied to complex and flexible bodies and tools by modeling the relationship between the body and tool using a neural network that can consider implicit grasping states.

  This study is organized as follows.
  In \secref{sec:proposed-method}, we describe the network structure of the Tool-Body Network with Parametric Bias (TBNPB), its training, online update of grasping state, and tool-tip position estimation and control.
  In \secref{sec:experiments}, we confirm the effectiveness of this study on the simulation of PR2, the actual PR2, and the musculoskeletal humanoid MusashiLarm.
  In \secref{sec:discussion}, we discuss the experimental results and conclude in \secref{sec:conclusion}.
}%
{%
  本研究の代替となり得る手法には主に, (1)visual/tactile feedbackを使った手法, (2)幾何モデルを使った把持状態推定手法が考えられる.
  (1)は, 道具使用におけるセンサ変化を記憶\cite{pastor2012associative}または学習\cite{girgin2018associative}しておくことによるunexpectedな把持状態変化に対してロバストに対応可能なtactile feedback, 道具先端位置に対するvisual feedbackが考えられる.
  (2)は, 道具の幾何モデルを使い, 手先の位置と道具先端位置の関係から道具の把持位置や角度を割り出すシンプルな方法である.
  (1)は把持状態を推定せずにセンサフィードバックによりそれらを補償する方法, (2)は幾何モデルから把持状態を理解して道具を用いる方法と言うことができる.
  しかし, (1)は操作したい状態と制御入力のJacobianが自明ではないdeformable toolや複雑なハンド・ロボット構造等には対応できない.
  また, 教示した動作への追従を主としており, 道具・把持状態のモデル化が行われないため, 本研究とは適用範囲が大きく異なる.
  この他にも, 模倣学習\cite{sasagawa2021autoregressive}によるsensor feedbackを使ったtool-tip controlもあるが, 道具の把持状態変化に適応した例はない.
  (2)は幾何モデルを前提としているためdeformable toolや複雑な把持状態等を扱うことができない.
  これらに対して, 本研究はニューラルネットワークによる身体と道具の関係のモデル化・暗黙的な把持状態の考慮により, 複雑で柔軟な身体・ハンド・道具等に適用可能な, 汎用的なモデルを提供する.

  本研究の構成は以下のようになっている.
  \secref{sec:proposed-method}ではParametric Biasを含むTool-Body Networkのネットワーク構成・学習・オンライン適応・道具先端位置の推定と制御について順に述べる.
  \secref{sec:experiments}では, PR2のSimulation, PR2, 筋骨格ヒューマノイドMusashiLarmにおいて本研究の有効性を確かめる.
  \secref{sec:discussion}では実験結果について議論し, \secref{sec:conclusion}で結論を述べる.
}%

\section{Tool-Body Network with Parametric Bias} \label{sec:proposed-method}
\switchlanguage%
{%
  In this study, we call the network representing the static relationship between a tool-tip and the body control command with parametric bias, Tool-Body Network with Parametric Bias (TBNPB).
  The overall system of this study surrounding TBNPB is shown in \figref{figure:whole-system}.
  First, the network structure of TBNPB is constructed (\secref{subsec:network-structure}), and TBNPB is trained offline (\secref{subsec:training}).
  Second, the grasping state is updated online through parametric bias (\secref{subsec:online-update}), and the tool-tip is estimated and controlled using TBNPB (\secref{subsec:estimation-control}).
}%
{%
  本研究では, 道具の先端と身体の制御入力間の間の静的な関係を表すParametric Biasを含んだネットワークを, Tool-Body Network with Parametric Bias (TBNPB)と呼ぶ.
  TBNPBを囲む本研究の全体システムを\figref{figure:whole-system}に示す.
}%

\subsection{Network Structure of TBNPB} \label{subsec:network-structure}
\switchlanguage%
{%
  The network structure of TBNPB is simple and can be expressed as follows,
  \begin{align}
    \bm{x}_{tool} = \bm{h}(\bm{u}, \bm{p})
  \end{align}
  where $\bm{x}_{tool}$ is the tool-tip position, $\bm{h}$ is TBNPB, $\bm{u}$ is the body control command, and $\bm{p}$ is the parametric bias, which corresponds to the implicit grasping state.
  Although $\bm{x}_{tool}$ can represent position and orientation, in this study, it represents only the three-dimensional position.
  In this study, $\bm{u}$ represents the control command of the joint angle $\bm{\theta}^{ref}$.
  Parametric bias $\bm{p}$ has been originally used to extract multiple attractor dynamics in time series information \cite{tani2004parametric}.
  Therefore, it is mostly used together with recurrent neural networks, but in this study, we use this parametric bias for the static correspondence network.

  In this study, the number of layers of TBNPB is 7.
  The number of units is set to the combined number of dimensions of $\bm{u}$ and $\bm{p}$ (which varies depending on the robot) for the input, 300 for all the middle 5 layers, and 3 (the number of dimension of $\bm{x}_{tool}$) for the output.
  The activation function is hyperbolic tangent, and the update rule is Adam \cite{kingma2015adam}.
  The input and output values of the network are normalized using the data obtained during training.
}%
{%
  TBNPBのネットワーク構造は非常に単純で, 以下のように表される.
  \begin{align}
    \bm{x}_{tool} = \bm{h}(\bm{u}, \bm{p})
  \end{align}
  ここで, $\bm{x}_{tool}$は道具の先端位置, $\bm{h}$はTBNPBのネットワーク, $\bm{u}$はロボット身体の制御入力, $\bm{p}$はParametric Biasとする.
  $\bm{x}_{tool}$は位置姿勢を表すことも可能だが, 本研究では3次元の位置とする.
  $\bm{u}$は本研究では関節角度の指令値$\bm{\theta}^{ref}$を表すこととする.
  Parametric Biasはもともと, 時系列の情報に置いて, 複数のattractor dynamicsを抽出するために用いられていた\cite{tani2004parametric}.
  そのためrecurrent neural networkと一緒に用いる場合がほとんどであるが, 本研究では静的な対応付けのネットワークにこのparametric biasを用いている.

  本研究では, ネットワーク構造は7層とし, ユニット数については, 入力は$\bm{u}$と$\bm{p}$を合わせた次元数(ロボットによって変わる), 中間の5層は全て300, 出力は$\bm{x}_{tool}$の3次元とした.
  活性化関数はtanh, 更新則はAdam \cite{kingma2015adam}とする.
  ネットワークの入力と出力は訓練時に得られたデータを使って正規化されている.
}%

\subsection{Training of TBNPB} \label{subsec:training}
\switchlanguage%
{%
  This section describes Data Collector and Network Trainer in \figref{figure:whole-system}.
  First, for various grasping states $k$ ($1 \leq k \leq K$; $K$ is the total number of grasping states used for training), where the grasped angle and position of the tool are different, the data at various body control commands $D_{k}=\{(\bm{u}, \bm{x}_{tool})_{1}, \cdots, (\bm{u}, \bm{x}_{tool})_{N_{k}}\}$ ($N_{k}$ is the number of data for grasping state $k$) is collected.
  Also, we prepare parametric bias $\bm{p}_k$ for each grasping state $k$ (all $\bm{p}_k$ are initialized to 0).
  Thus, the data $D_{train} = \{(D_{1}, \bm{p}_{1}), \cdots, (D_{N_{k}}, \bm{p}_{N_k})\}$ is collected and it is used to train $\bm{h}$.
  Here, $\bm{p}_{k}$ is common for the data $D_{k}$ and different variables are used for different grasping states.
  During the training process, the network weights $W$ and $\bm{p}_k$ are updated at the same time by the backpropagation method.
  In this way, the grasping state information is embedded in $\bm{p}_{k}$.
  No annotation for $\bm{p}_{k}$ is necessary.

  In this study, training procedure is performed in two stages.
  First, we collect data by changing the grasping state in the simulation, and calculate $W$ and $\bm{p}_{k}$.
  Then, we initialize $\bm{p}_{k}$ to 0, leaving only the $W$ calculated in the simulation.
  Finally, we collect the data in the actual robot and perform the training again.
  Since the data obtained from the actual robot is small, we conduct fine-tuning.
}%
{%
  本項は\figref{figure:whole-system}のData Collector, Network Trainerの説明である.
  まず, 道具の持つ角度や持つ位置が異なるような, 様々な把持状態$k$ ($1 \leq k \leq K$, $K$は訓練に使う全把持状態数)について, 様々な姿勢におけるデータ$D_{k}=\{(\bm{u}, \bm{x}_{tool})_{1}, \cdots, (\bm{u}, \bm{x}_{tool})_{N_{k}}\}$を収集する($N_{k}$は把持状態$k$に関するデータ数).
  また, それぞれの把持状態$k$についてparametric bias $\bm{p}_k$を用意する(全ての$\bm{p}_k$は0に初期化されている).
  よって, データ$D_{train} = \{(D_{1}, \bm{p}_{1}), \cdots, (D_{N_{k}}, \bm{p}_{N_k})\}$が収集され, この$D_{train}$を用いて$\bm{h}$を学習させる.
  このとき, $\bm{p}_{k}$はデータ$D_{k}$については共通であり, 異なる把持状態については異なる変数とする.
  学習の際はネットワークの重み$W$と同時に$\bm{p}_k$も誤差逆伝播法によって更新する.
  これにより, $\bm{p}_{k}$に把持状態に関する情報が埋め込まれる.

  本研究では学習は2段階で行う.
  まず, シミュレーション上で把持位置や把持姿勢を変化させてデータを収集し, $W$と$\bm{p}_{k}$を計算する.
  その後, シミュレーションで計算された$W$のみ残し, $\bm{p}_{k}$を0に初期化する.
  最後に, 実機においてデータを収集し, 学習を行う.
  実機で得られるデータは少数なため, fine-tuningにより対応する.
}%

\subsection{Online Update of Grasping State} \label{subsec:online-update}
\switchlanguage%
{%
  This section describes Grasping State Updater in \figref{figure:whole-system}.
  Assuming that the grasping state can change at any time, we update the parametric bias $\bm{p}$ online.
  Data is collected when the tool-tip position $\bm{x}_{tool}$ is recognized and the control command $\bm{u}$ differs to a certain extent from the control command $\bm{u}^{prev}$ collected just before; that is, if $||\bm{u}-\bm{u}^{prev}||_{2}>C_{collect}$ ($||\cdot||_{2}$ is the L2 norm and $C_{collect}$ is the threshold).
  We start online update when the number of obtained data $N^{online}$ exceeds $N^{online}_{thre}$, and then we update $\bm{p}$ each time new data is collected.
  The weight $W$ is fixed; only $\bm{p}$ is updated as $N^{online}_{batch}$ batches and $N^{online}_{epoch}$ epochs.
  Here, the update rule is momentum SGD \cite{qian1999momentum} with the learning rate set to 0.1.
  The maximum number of the data is set to $N^{online}_{max}$ ($N^{online}_{thre} \leq N^{online}_{max}$), and the data exceeding it are deleted from the oldest one.
  By fixing the weight $W$ of the network and updating only the parametric bias, which has a small dimension, we can update only the grasping state while preventing over-fitting.

  In this study, we set $C_{collect}=10.0$ [deg], $N^{online}_{thre}=10$, $N^{online}_{batch}=N^{online}$, $N^{online}_{epoch}=3$, and $N^{online}_{max}=20$.
  Also, the sampling rate of data collection is 5 Hz.
  $C_{collect}$ should be set appropriately according to the scale of the whole motion.
  The larger $N^{online}_{thre}$ is, the more stable the online update is in the early stage, but the slower the update starts, so it should be set appropriately according to the application.
  The larger $N^{online}_{max}$ is, the more accurately the grasping state can be updated using a large number of data, but the slower it is to adapt to changes in the grasping state, so it should be set appropriately taking into account the tradeoff.
}%
{%
  本項は\figref{figure:whole-system}のGrasping State Updaterの説明である.
  把持状態が常に変化しうることを想定し, オンラインでparametric bias $\bm{p}$を更新する.
  道具の先端位置$\bm{x}_{tool}$が認識できているかつ, 制御入力$\bm{u}$が直前に収集された制御入力$\bm{u}^{prev}$と離れている, つまり$||\bm{u}-\bm{u}^{prev}||_{2}>C_{collect}$の場合に, データを収集する($||\cdot||_{2}$はL2ノルム, $C_{collect}$は閾値とする).
  得られたデータ数$N^{online}$が, $N^{online}_{thre}$を超えてから学習を開始し, その後新しいデータ収集される度に学習を行う.
  重み$W$を固定し, $\bm{p}$のみをバッチ数$N^{online}_{batch}$, エポック数を$N^{online}_{epoch}$として更新する.
  この際の更新則は, 学習率を0.1としたmomentum SGDとする.
  データは$N^{online}_{max}$ ($N^{online}_{thre} \leq N^{online}_{max}$)を最大値とし, それを超えたデータは古いものから順に削除していく($N^{online} \leq N^{online}_{max}$).
  ネットワークの重み$W$を固定し, 小さな次元であるparametric biasのみ更新することで, 過学習を防ぎつつ把持状態のみを更新することができる.

  本研究では, $C_{collect}=10.0$ [deg], $N^{online}_{thre}=10$, $N^{online}_{batch}=N^{online}$, $N^{online}_{epoch}=3$, $N^{online}_{max}=20$とした.
  また, data collectionのsampling rateは5 Hzである.
  $C_{collect}$は動作全体のスケールに応じて適切に設定すべきである.
  $N^{online}_{thre}$は大きい方が学習初期は安定するが, その分学習し始めるのが遅くなり, 基本的には大きめに設定すべきである.
  $N^{online}_{max}$は大きいほど多くのデータを使って正確にgrasping stateを更新できるものの, 把持状態変化への適応には遅くなるため, そのtradeoffを加味して適切に設定すべきである.
}%

\subsection{State Estimation and Control of Tool-Tip Using TBNPB} \label{subsec:estimation-control}
\switchlanguage%
{%
  This section describes Tool-Tip State Estimator / Controller in \figref{figure:whole-system}.
  The tool-tip state estimation is very simple and it can be calculated by merely inputting the current $\bm{p}$ and the control command $\bm{u}$ into $\bm{h}$.
  The tool-tip control is performed by optimization using the backpropagation method and gradient descent.
  First, we obtain the current control command $\bm{u}^{cur}$ and use it as the initial value of the control command $\bm{u}^{opt}$ to be optimized.
  Next, we perform the optimization as follows,
  \begin{align}
    \bm{x}^{est}_{tool} &= \bm{h}(\bm{u}^{opt}, \bm{p})\\
    L &= ||\bm{x}^{est}_{tool}-\bm{x}^{ref}_{tool}||_{2} + \alpha{L}_{const}(\bm{u}^{opt})\\
    \bm{u}^{opt} &\gets \bm{u}^{opt} + \gamma\partial{L}/\partial{\bm{u}^{opt}}
  \end{align}
  where $\bm{x}^{ref}_{tool}$ is the target tool-tip position, $L_{const}$ is the constraint on the control command $\bm{u}$, $\alpha$ is the weight of the loss function, and $\gamma$ is the learning rate.
  For $L_{const}$, for example, if we want $\bm{u}$ to be as close as possible to the current control command, we can set it to $||\bm{u}^{opt}-\bm{u}^{cur}||_{2}$, or if we do not want to move a certain joint, we can constrain it by giving a loss function only for that joint.
  For $\gamma$, in this study, we try $\bm{N}^{control}_{batch}$ number of $\gamma$ from 0 to $\gamma^{max}$ by line search and adopt the one with the smallest loss, repeating $N^{control}_{epoch}$ times ($\gamma^{max}$ is the maximum value of $\gamma$).
  Using the finally obtained $\bm{u}^{opt}$, the tool-tip position can be controlled.

  In this study, we set $\gamma^{max}=0.5$, $N^{control}_{batch}=30$, and $N^{control}_{epoch}=10$.
}%
{%
  本項は\figref{figure:whole-system}のTool-Tip State Estimator / Controllerの説明である.
  道具先端位置の推定は非常に単純であり, 現在の$\bm{p}$と制御入力$\bm{u}$を$\bm{h}$に入力するのみで計算することができる.
  道具先端位置の制御は誤差逆伝播法と勾配降下を使った最適化により行う.
  まず, 現在の制御入力$\bm{u}^{cur}$を取得し, これを最適化する制御入力$\bm{u}^{opt}$の初期値とする.
  次に, 以下のように最適化を行う.
  \begin{align}
    \bm{x}^{est}_{tool} &= \bm{h}(\bm{u}^{opt}, \bm{p})\\
    L &= ||\bm{x}^{est}_{tool}-\bm{x}^{ref}_{tool}||_{2} + \alpha\bm{L}_{const}(\bm{u}^{opt})\\
    \bm{u}^{opt} &\gets \bm{u}^{opt} + \gamma\partial{L}/\partial{\bm{u}^{opt}}
  \end{align}
  ここで, $\bm{x}^{ref}_{tool}$は道具先端位置の指令値, $L_{const}$は制御入力$\bm{u}$に関する制約を表し, $\alpha$は損失関数の重み, $\gamma$は学習率を表す.
  $L_{const}$については, 例えば$\bm{u}$を現在の制御入力になるべく近いかたちにしたければ$||\bm{u}^{opt}-\bm{u}^{cur}||_{2}$にしたり, ある一つの関節を動かしたくなければ, その関節のみに対して損失を与えて制約をかけたりすることが可能である.
  $\gamma$は, 本研究ではline searchによって, 0から$\gamma^{max}$までの$\bm{N}^{control}_{batch}$種類の$\gamma$を試し, 最も$L$が小さかったものを採用することを$N^{control}_{epoch}$回繰り返す($\gamma^{max}$は$\gamma$の最大値とする).
  最終的に得られた$\bm{u}^{opt}$を使い, 道具の先端位置を制御することができる.

  本研究では, $\gamma^{max}=0.5$, $N^{control}_{batch}=30$, $N^{control}_{epoch}=10$とする.
}%

\begin{figure}[t]
  \centering
  \includegraphics[width=0.8\columnwidth]{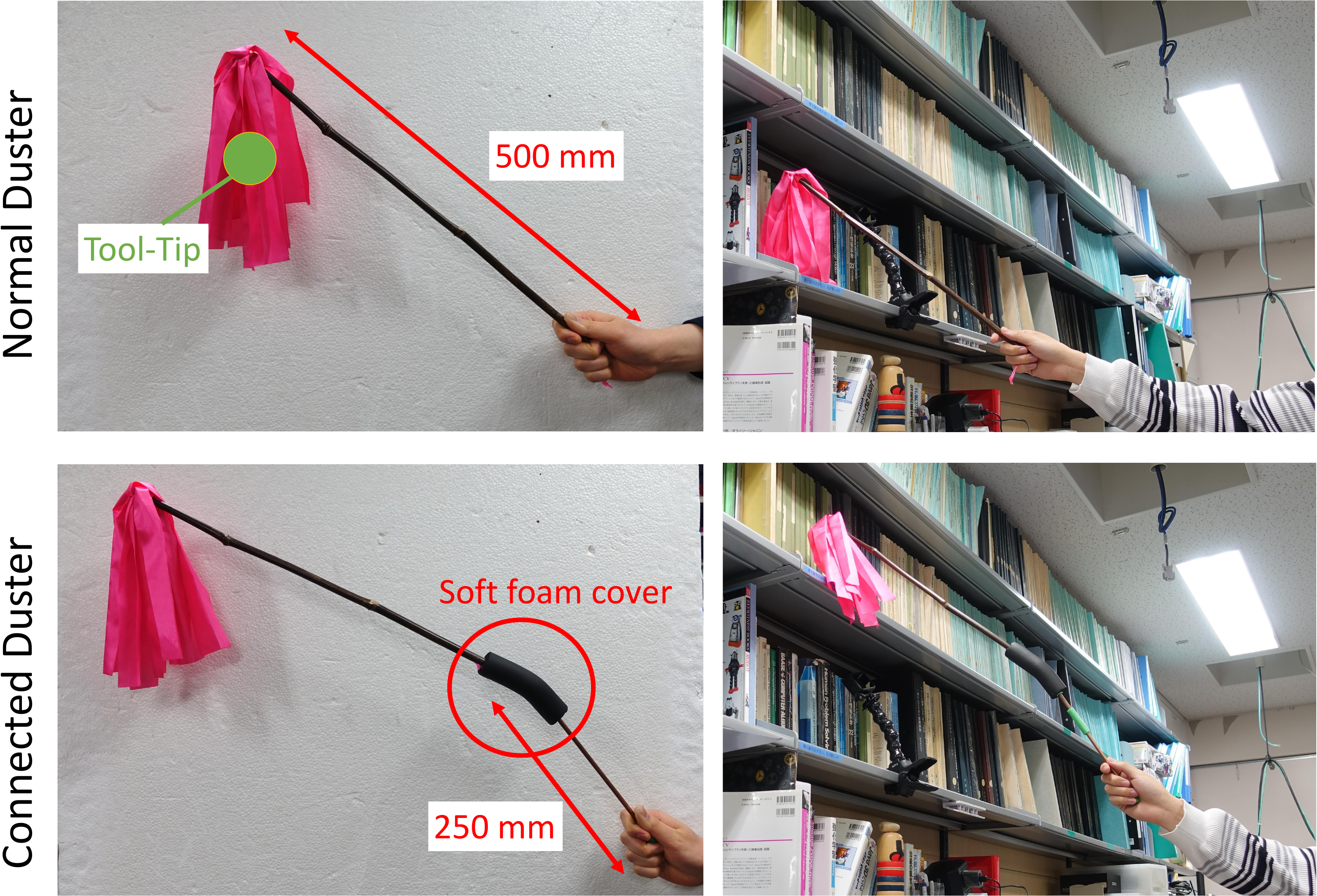}
  \vspace{-1.0ex}
  \caption{The tools used in this study: normal and connected dusters.}
  \label{figure:hataki-setup}
  \vspace{-1.0ex}
\end{figure}

\begin{figure}[t]
  \centering
  \includegraphics[width=0.8\columnwidth]{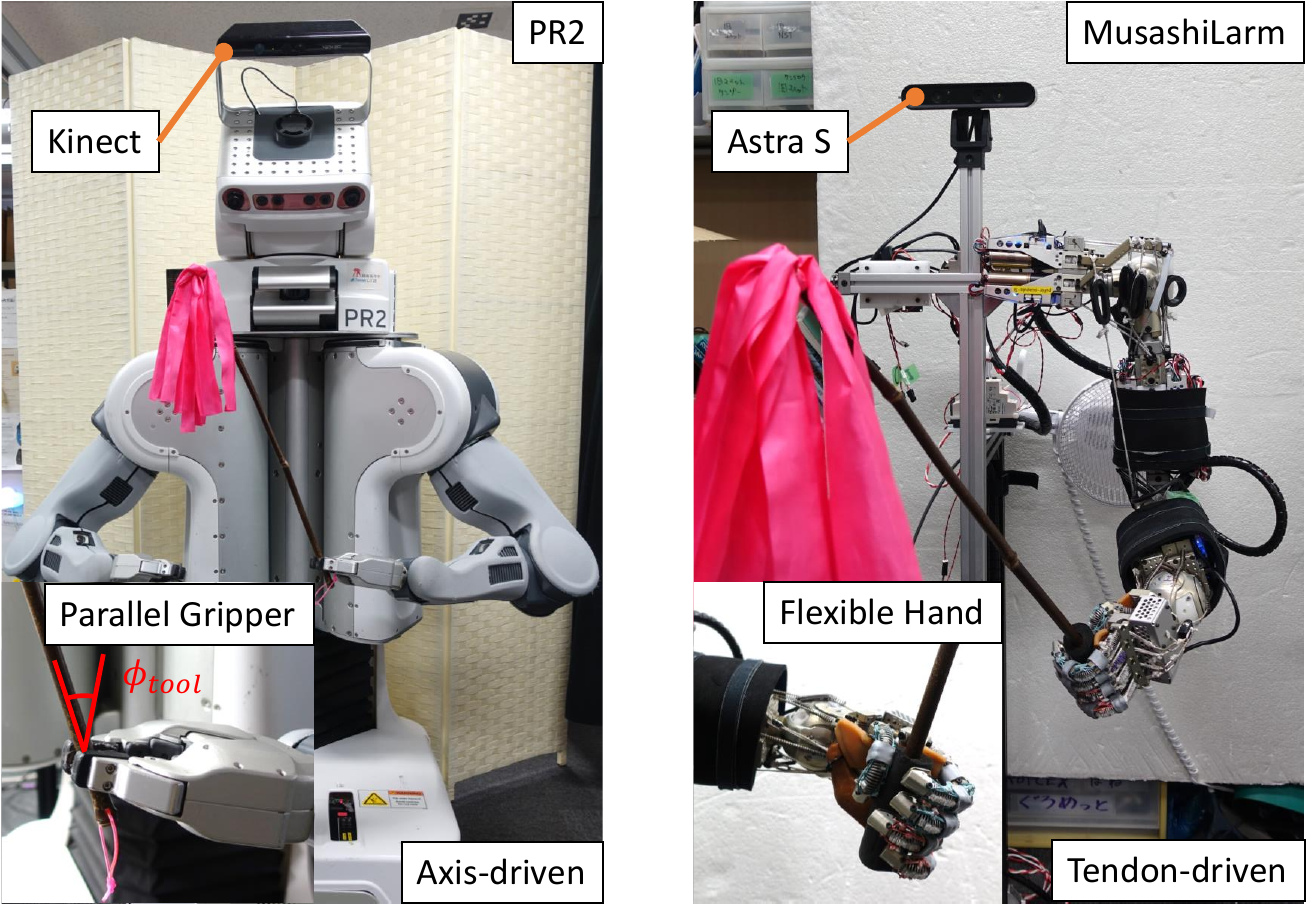}
  \vspace{-1.0ex}
  \caption{The robots used in this study: PR2 with the parallel gripper and the musculoskeletal arm MusashiLarm with the flexible hand.}
  \label{figure:robot-setup}
  \vspace{-1.0ex}
\end{figure}

\section{Experiments} \label{sec:experiments}
\subsection{Experimental Setup} \label{subsec:experimental-setup}
\switchlanguage%
{%
  In this study, we conduct experiments using a duster, which removes dust from shelves and objects by controlling the tool-tip position (\figref{figure:hataki-setup}).
  A colored cloth is attached to the tool-tip, and the tool-tip position is recognized by extracting the color of the cloth.
  The cloth of the duster drops from the tip of the stick in the direction of gravity, and the tool-tip position cannot be linearly transformed from the hand posture.
  As a more difficult condition, we also use another duster of which the length is increased by attaching an additional stick to it in the PR2 experiment.
  We call this a ``connected duster''.
  The normal duster and the additional stick are connected by a flexible foam cover, so that the tool-tip position changes greatly depending on the angle at which the duster is held.
  The stick length of the normal duster is 500 mm, that of the colored cloth is 200 mm, and that of the additional stick is 250 mm.

  In the experiments of this study, we use the simulation and the actual robot of the wheeled axis-driven humanoid PR2 and the actual robot of the musculoskeletal arm MusashiLarm \cite{kawaharazuka2019musashi} (\figref{figure:robot-setup}).
  In the PR2 / MusashiLarm, the head is equipped with a Kinect (Microsoft, Corp.) / Astra S (Orbbec 3D Technology International, Inc.) depth camera.
  Point clouds of the tool-tip are extracted through color filtering, euclidean clustering is performed, and the center of the largest cluster is set as the tool-tip position.
  The hand of PR2 is a parallel gripper, and the grasping angle $\phi_{tool}$ shown in \figref{figure:robot-setup} and the grasping position of the tool are mainly changed during the tool-use.
  On the other hand, the hand of MusashiLarm is a flexible hand using machined springs, and it is difficult to parameterize the grasping state.
}%
{%
  本研究では, 掃除用具の一つであるはたきを使った実験を行う(\figref{figure:hataki-setup}).
  はたきはその道具の先端位置を制御することで, 棚や隙間のほこりを落とす道具である.
  はたきの先端には色のついた布が付いており, 道具の先端位置はこの色を抽出することで認識する.
  このはたきの布は持つ角度によって棒の先端から重力方向に下がり, 手先位置姿勢から道具先端位置をアフィン変換することはできない.
  また, より難しい問題として, 追加の棒を括りつけることで長さを増したはたきもPR2実験において利用する.
  はたきと追加の棒は柔軟な発泡性の素材により接続し, はたきを持つ角度によって道具先端位置が大きく変化するようになっている.
  なお, 本研究で使うはたきは棒の長さが500 mm, 布は200 mm, 追加の棒は250 mmであり, 通常のはたきをnormal, 追加の棒により接続された長いはたきをconnectedと呼ぶ.

  本研究の実験では, 台車型ロボットPR2のシミュレーションと実機, 筋骨格ヒューマノイドMusashiLarm \cite{kawaharazuka2019musashi}の実機を用いて実験を行う(\figref{figure:robot-setup}).
  PR2 / MusashiLarmにおいては, 頭部にKinect (Miscrosoft, Corp.) / Astra S (Orbbec 3D Technology International, Inc.)のdepth cameraがついており, 色抽出した道具の先端の点群をeuclidean clusteringし, 一番大きなclusterの中心位置を道具の先端位置とした.
  PR2のハンドは平行グリッパ型であり, はたきの把持の際は, 主に平行グリッパと垂直方向に関する角度$\phi_{tool}$, 道具を持つ位置, が変化する.
  これに対して, MusashiLarmのハンドは切削ばねを使った柔軟ハンドであり, 把持状態をパラメータ化することは難しい.
}%

\begin{figure}[t]
  \centering
  \includegraphics[width=0.8\columnwidth]{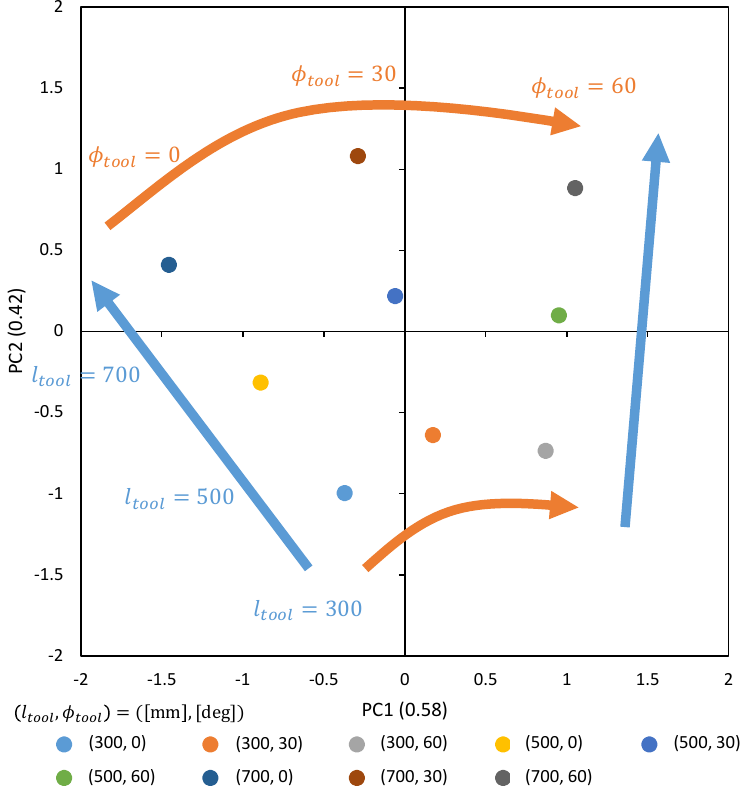}
  \vspace{-1.0ex}
  \caption{Parametric bias trained in PR2 simulation experiment.}
  \label{figure:sim-pb-exp}
  \vspace{-1.0ex}
\end{figure}

\begin{figure}[t]
  \centering
  \includegraphics[width=1.0\columnwidth]{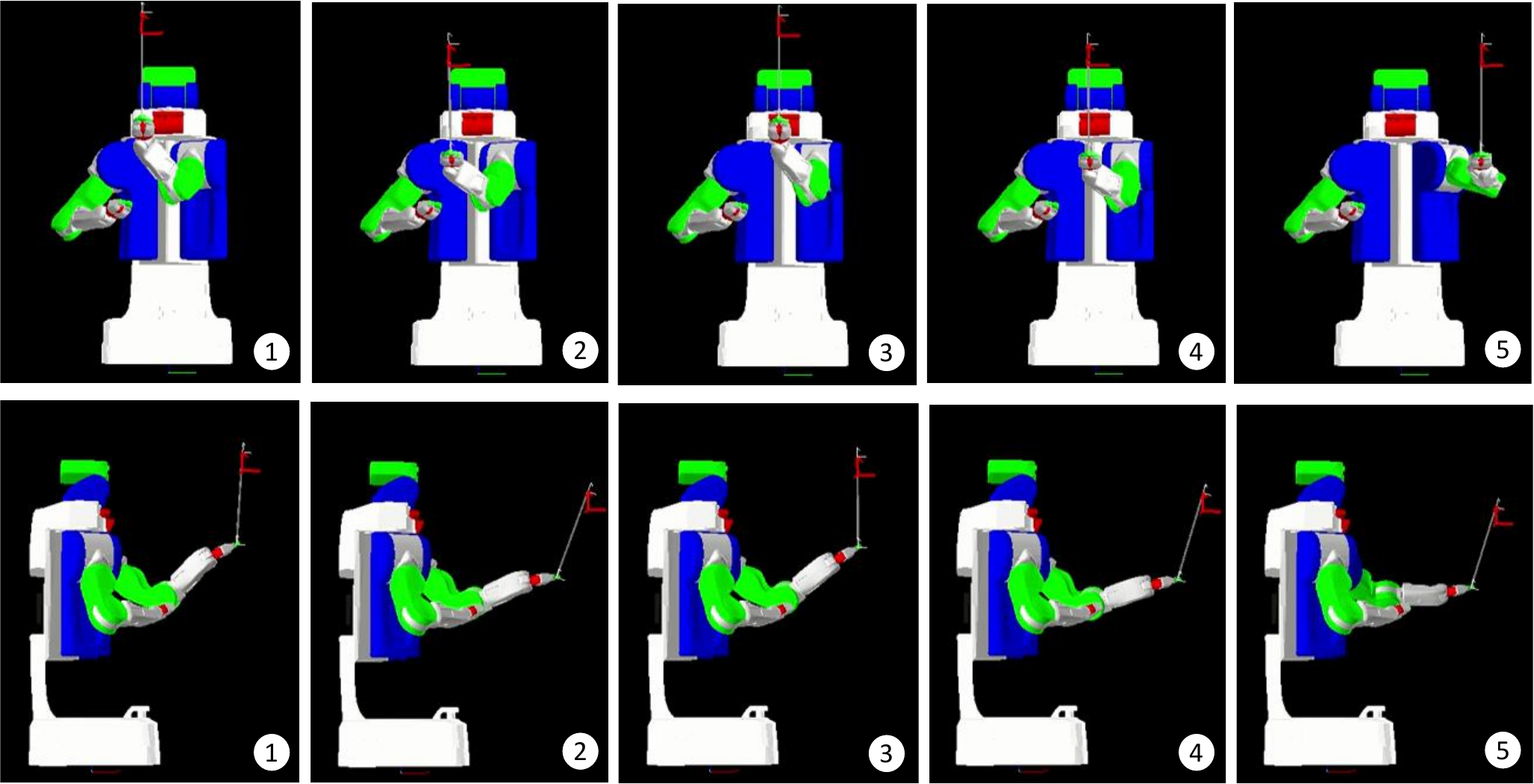}
  \vspace{-3.0ex}
  \caption{Duster-use motion for PR2 experiments.}
  \label{figure:sim-hataki-exp}
  \vspace{-3.0ex}
\end{figure}

\begin{figure}[t]
  \centering
  \includegraphics[width=0.8\columnwidth]{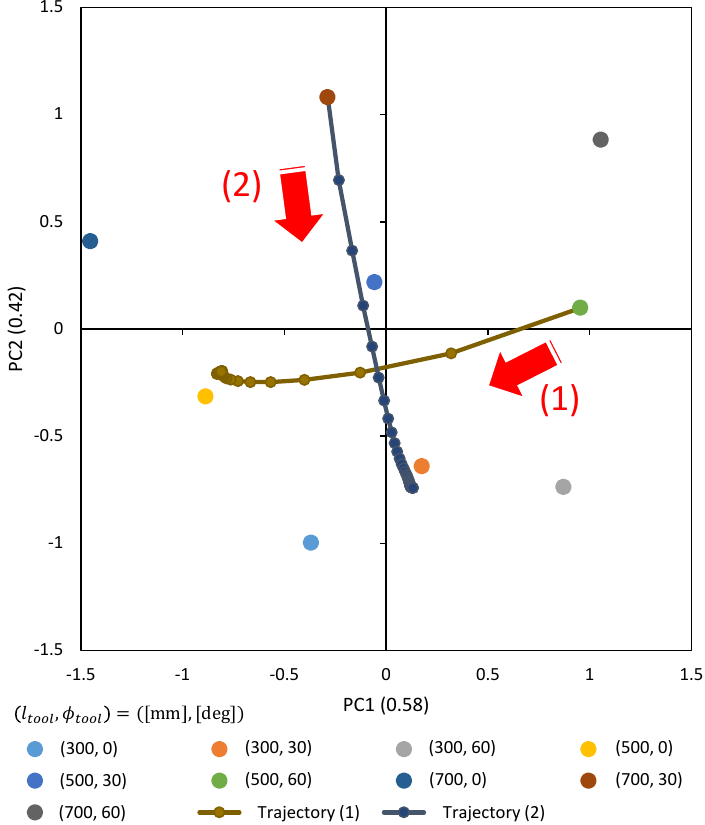}
  \vspace{-1.0ex}
  \caption{Transition of parametric bias when changing grasping state: (1) $(l_{tool}, \phi_{tool})=(500, 60) \rightarrow (500, 0)$, (2) $(l_{tool}, \phi_{tool})=(700, 30) \rightarrow (300, 30)$ in PR2 simulation experiment with grasping state updater.}
  \label{figure:sim-online-pb-exp}
  \vspace{-1.0ex}
\end{figure}

\begin{figure}[t]
  \centering
  \includegraphics[width=0.9\columnwidth]{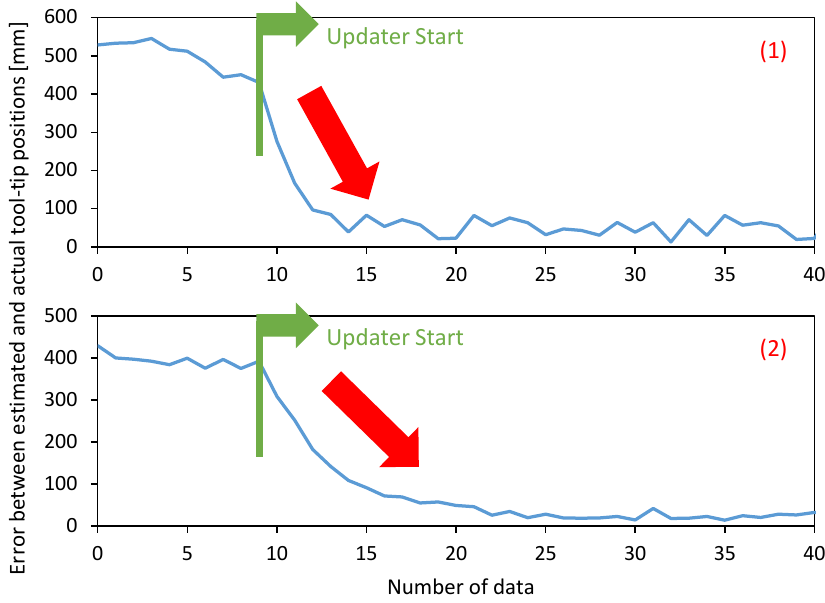}
  \vspace{-1.0ex}
  \caption{Transition of state estimation error of tool-tip position when changing grasping state: (1) $(l_{tool}, \phi_{tool})=(500, 60) \rightarrow (500, 0)$, (2) $(l_{tool}, \phi_{tool})=(700, 30) \rightarrow (300, 30)$ in PR2 simulation experiment with grasping state updater.}
  \label{figure:sim-online-est-exp}
  \vspace{-3.0ex}
\end{figure}

\begin{figure}[t]
  \centering
  \includegraphics[width=1.0\columnwidth]{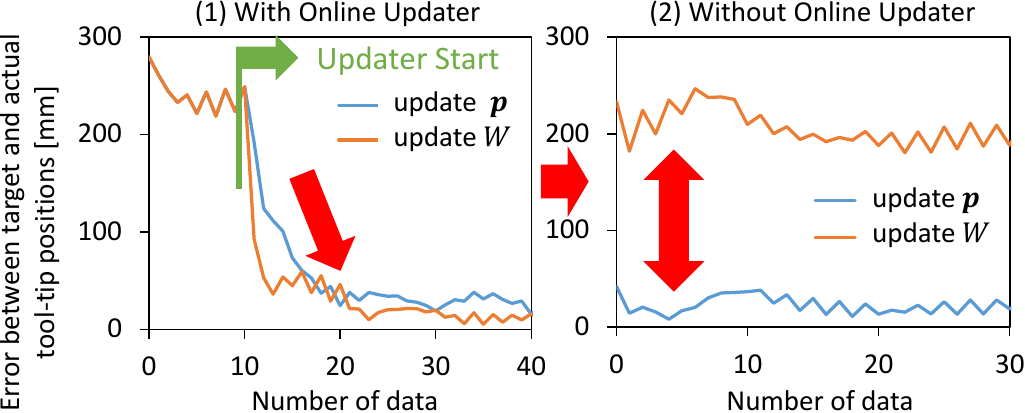}
  \vspace{-3.0ex}
  \caption{Transition of control error of tool-tip position when changing grasping state: $(l_{tool}, \phi_{tool})=(500, 30) \rightarrow (500, 60)$ in PR2 simulation experiment with online updaters of updating $\bm{p}$ or $W$. The right experiment is conducted at different posture from the trained one without online updaters.}
  \label{figure:sim-control-exp}
  \vspace{-3.0ex}
\end{figure}

\subsection{PR2 Simulation Experiment} \label{subsec:simulation-exp}
\switchlanguage%
{%
  We conduct an experiment using the geometric simulator of PR2.
  First, we attach a long thin object to the hand to represent the duster, and obtain data by changing the grasping position (expressed as the length of tool stick from the hand) $l_{tool}$ and the grasping angle (expressed as the angle perpendicular to the parallel gripper with one degree of freedom) $\phi_{tool}$.
  Since the cloth of the duster hangs down from the tip of the stick in the direction of gravity, we simulate the tool-tip at -100 mm in the $z$-direction from the tip of the stick.
  We change grasping state as $l_{tool}=\{300, 500, 700\}$ [mm] and $\phi_{tool}=\{0, 30, 60\}$ [deg].
  Next, the joint angle limit is determined, and within the range, the joint angles are randomly sampled for each grasping state, and the data $D_{train}$ is obtained.
  The total number of $D_{train}$ is 9000, and TBNPB is trained as 300 batches and 300 epochs.
  Note that $\bm{u}$ is seven-dimensional and $\bm{p}$ is two-dimensional.
  The parametric bias $p_k$ obtained here is represented in two-dimensional space through principal component analysis (PCA) as shown in \figref{figure:sim-pb-exp}.
  We can see that the parametric bias is aligned neatly along the magnitude of $l_{tool}$ and $\phi_{tool}$.
  The larger $l_{tool}$ is, the larger the difference in parametric bias due to the change in $\phi_{tool}$ is, which is consistent with the fact that a longer tool has a larger change in tool-tip position depending on the angle.

  Next, we experiment on the behavior of grasping state updater and tool-tip state estimator.
  We conduct experiments for two cases: (1) when the grasping state is changed from $(l_{tool}, \phi_{tool})=(500, 60)$ to $(l_{tool}, \phi_{tool})=(500, 0)$, and (2) when the grasping state is changed from $(l_{tool}, \phi_{tool})=(700, 30)$ to $(l_{tool}, \phi_{tool})=(300, 30)$.
  The motion of shaking the duster is shown in \figref{figure:sim-hataki-exp} (for the case of $(l_{tool}, \phi_{tool})=(500, 30)$).
  This is a motion in which we determine a reference point of the tool-tip (with the center of wheeled cart of PR2 as the origin, e.g. (800, -100, 1600) [mm] for $(l_{tool}, \phi_{tool})=(500, 30)$), and alternately move the tool-tip by 100 mm in the $y$ direction, then move and return (200, -200) [mm] in the $(x, z)$ direction, while solving inverse kinematics.
  If the movement in the $y$ direction exceeds 500 mm, move in the opposite direction.
  The transition of parametric bias during this motion is shown in \figref{figure:sim-online-pb-exp}, and the transition of the state estimation error of the tool-tip position is shown in \figref{figure:sim-online-est-exp}.
  It can be seen that for both (1) and (2), the parametric bias is gradually approaching the area around the current grasping state obtained at training.
  In addition, the state estimation error of the tool-tip position also decreases gradually.
  When more than 20 data points were collected, the average estimation errors were 52.2 mm for (1) and 25.9 mm for (2).

  Finally, we experiment on the tool-tip controller.
  Starting from the state where parametric bias is $(l_{tool}, \phi_{tool})=(500, 30)$ obtained at training, and setting $(l_{tool}, \phi_{tool})=(500, 60)$, we compare the control error of the tool-tip position when using the grasping state updater (update $\bm{p}$) or when $\bm{p}$ is fixed and $W$ is updated (update $W$).
  The former corresponds to updating only $\bm{p}$, while the latter corresponds to updating the weight $W$ without $\bm{p}$, as in ordinary online learning (note that the learning rate is set to 0.01 for the latter).
  The behavior is the same as that of \figref{figure:sim-hataki-exp}.
  Here, the joint angle $\bm{u}^{orig}$ of \figref{figure:sim-hataki-exp} generated as $(l_{tool}, \phi_{tool})=(500, 30)$ is used as a reference, so that we set $\alpha=0.3$ and $L_{const}=||\bm{u}^{opt}-\bm{u}^{orig}||_{2}$.
  The transition of the control error of the tool-tip position is shown in the left figure of \figref{figure:sim-control-exp}.
  It can be seen that the initial control error without online updaters is about 240 mm, while the control error is greatly reduced by online updaters.
  For more than 20 data points, the average control error is 31.5 mm for the former (update $\bm{p}$) and 19.2 mm for the latter (update $W$), and the latter, which updates the entire network, is more accurate.
  The transition of the control error when the online updater was stopped and the same tool-tip position trajectory was performed with completely different $\bm{u}^{orig}$ due to different tool-tip rotational constraints is shown in the right figure of \figref{figure:sim-control-exp}.
  After updating only $\bm{p}$, the control error is 22.6 mm on average, while it is 207 mm after updating $W$.
  In the case of updating only $\bm{p}$, the grasping state updater is effective in other joint angles, while in the case of updating $W$, the control error is larger in other joint angles due to over-fitting to the data used for online learning.
}%
{%
  PR2の幾何シミュレータを使って実験を行う.
  まず, 擬似的な真っ直ぐな棒をはたきとして用意し, これを把持位置(手先からの道具の長さで表現)$l_{tool}$, 把持角度(平行グリッパと垂直方向の角度を一自由度で表現)$\phi_{tool}$を3種類ずつ変化させながらデータを取得する.
  はたきの布は棒の先端から重力方向に垂れるため, 道具先端位置は棒の先端からz方向に-100 mmの場所としてシミュレーションを行う.
  $l_{tool}=\{300, 500, 700\}$ [mm], $\phi_{tool}=\{0, 30, 60\}$ [deg]とした.
  次に, 関節角度範囲を決め, その中で, それぞれの把持状態について関節角度をランダムにサンプリングし, データ$D_{train}$を取得する.
  得られた$D_{train}$は合計で9000個であり, これをバッチ数300, エポック数300として学習させる.
  なお, $\bm{u}$は7次元であり, $\bm{p}$は2次元とした.
  このときに得られたparametric bias $p_k$をPCAを通して2次元空間に表現した図を\figref{figure:sim-pb-exp}に示す.
  $l_{tool}$と$\phi_{tool}$の大小に伴って, それぞれのPBが綺麗に整列しているのがわかる.
  $l_{tool}$が大きいほど$\phi_{tool}$の変化によるPBの違いが大きくなっており, より長い道具ほど角度によって大きく道具先端位置が変化する点と一致している.

  次に, parametric biasのオンライン学習と道具先端位置推定の挙動について実験する.
  道具を(1) $(l_{tool}, \phi_{tool})=(500, 60)$から$(l_{tool}, \phi_{tool})=(500, 0)$に変化させた場合, (2) $(l_{tool}, \phi_{tool})=(700, 30)$から$(l_{tool}, \phi_{tool})=(300, 30)$に変化させた場合の2つの場合について実験を行う.
  はたきを振る動作は, \figref{figure:sim-hataki-exp}に示すようなものである($(l_{tool}, \phi_{tool})=(500, 30)$の道具を持った場合).
  道具先端位置の基準点を決め(PR2のベース中心を原点とする, $(l_{tool}, \phi_{tool})=(500, 30)$であれば例えば(800, -100, 1600) [mm]), y方向に一回100 mm進み, (x, z)方向に一回(200, -200) [mm]逆運動学を解いて動かして戻す, を交互に行う動作である.
  y方向には, 500 mm進んだら100mmずつ戻る.
  この動きの際のparametric biasの動きを\figref{figure:sim-online-pb-exp}に, 道具先端位置の推定誤差の遷移を\figref{figure:sim-online-est-exp}に示す.
  (1), (2)の両者について, parametric biasは訓練時に得られた現在の把持状態周辺に徐々に近づいていることがわかる.
  また, それに伴い, 道具先端位置の推定誤差も徐々に下がっていくことがわかる.
  データが20以上集まった時における推定誤差の平均はは(1)が52.2 mm, (2)は25.9 mmであった.

  最後に, 道具先端位置の制御について実験する
  parametric biasが訓練時に得られた$(l_{tool}, \phi_{tool})=(500, 30)$の状態から始め, $(l_{tool}, \phi_{tool})=(500, 60)$としたときに, grasping state updaterを使った場合(update $\bm{p}$)と, $\bm{p}$は固定して$W$を更新した場合(update $W$)での制御誤差を比較する.
  前者は$\bm{p}$のみを更新するのに対して, 後者は通常のオンライン学習のように, $\bm{p}$は存在せずに重み$W$を更新していくことに相当する(なお, 後者は学習率を0.01とした).
  動作は\figref{figure:sim-hataki-exp}と同じ動作である.
  このとき, $(l_{tool}, \phi_{tool})=(500, 30)$として生成された\figref{figure:sim-hataki-exp}の関節角度$\bm{u}^{orig}$を基準とするため, \secref{subsec:estimation-control}において$\alpha=0.3$, $L_{const}=||\bm{u}^{opt}-\bm{u}^{orig}||_{2}$とする.
  道具先端位置の制御誤差の遷移を\figref{figure:sim-control-exp}の左図に示す.
  online updaterが働かない初期は制御誤差が約240 mmなのに対して, online updaterによって制御誤差が大きく下がっていることがわかる.
  得られたデータ数が20以上について, 制御誤差の平均は, 前者で31.5 mm, 後者で19.2 mmであり, ネットワーク全体を更新する後者の方が精度が良かった.
  その後online updaterを止め, 道具先端の回転制約を変えて全く異なる$\bm{u}^{orig}$で同じ道具先端位置軌道の動作を行った際の制御誤差の遷移を\figref{figure:sim-control-exp}の右図に示す.
  $\bm{p}$のみ更新した後では, 制御誤差は平均で22.6 mmなのに対して, $W$を更新した後では207 mmであった.
  $\bm{p}$のみを更新した場合は他の関節角度においても把持状態更新が効果を発揮するのに対して, $W$を更新した場合は訓練に使ったデータに過学習してしまい, 他の関節角度においては制御誤差が大きくなってしまうことがわかった.
}%

\begin{figure}[t]
  \centering
  \includegraphics[width=0.8\columnwidth]{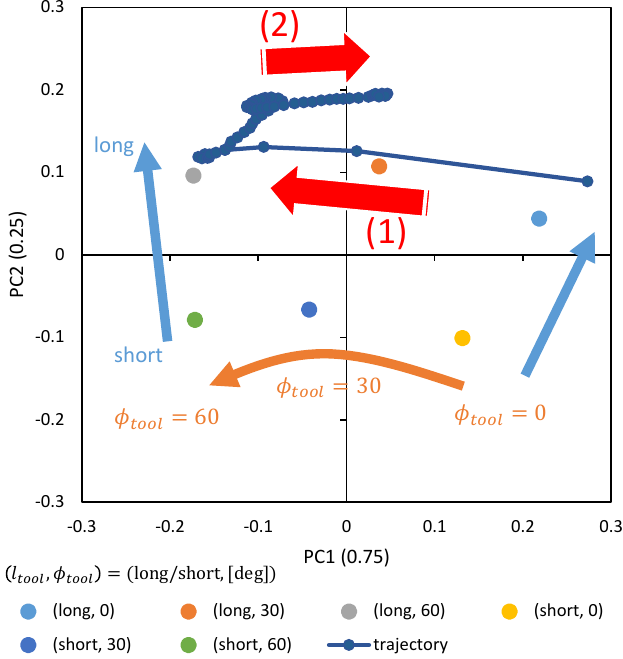}
  \vspace{-1.0ex}
  \caption{Parametric bias trained in PR2 experiment with the normal duster and its trajectory in the tool-tip control experiment with grasping state updater.}
  \label{figure:pr2-pb-exp}
  \vspace{-1.0ex}
\end{figure}

\begin{figure}[t]
  \centering
  \includegraphics[width=0.9\columnwidth]{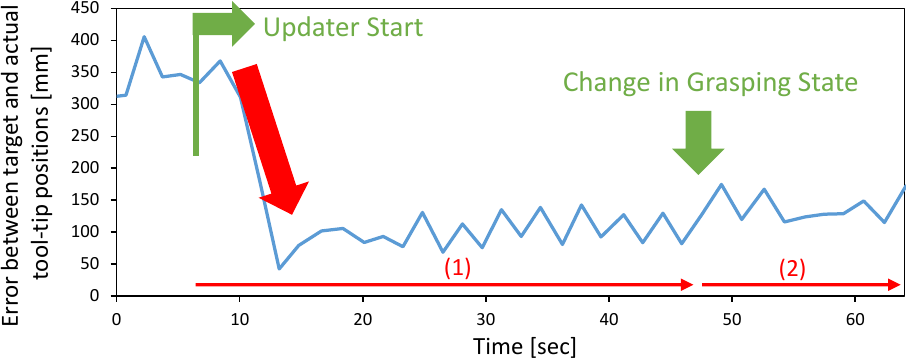}
  \vspace{-1.0ex}
  \caption{Transition of control error of tool-tip position in tool-tip control experiment of PR2 with the normal duster.}
  \label{figure:pr2-control-exp}
  \vspace{-3.0ex}
\end{figure}

\begin{figure}[t]
  \centering
  \includegraphics[width=0.8\columnwidth]{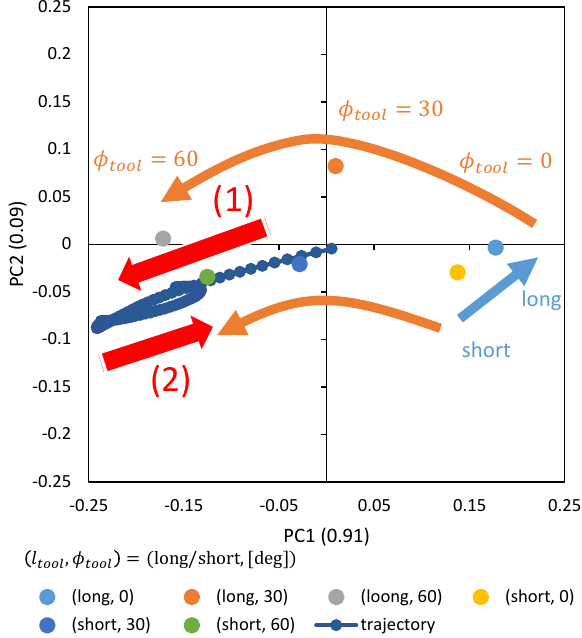}
  \vspace{-1.0ex}
  \caption{Parametric bias trained in PR2 experiment with the connected duster and its trajectory in the tool-tip control experiment with grasping state updater.}
  \label{figure:pr2-connected-pb-exp}
  \vspace{-1.0ex}
\end{figure}

\begin{figure}[t]
  \centering
  \includegraphics[width=0.9\columnwidth]{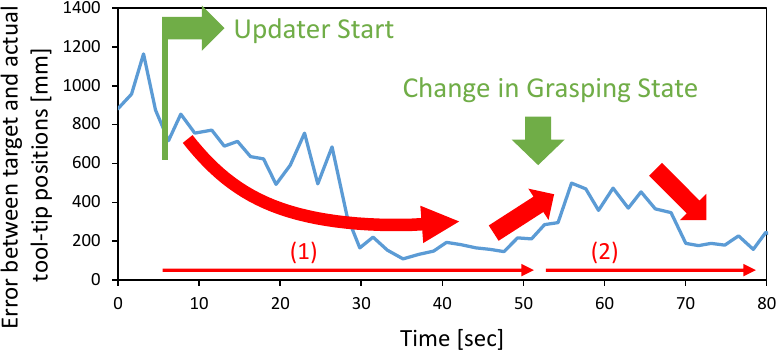}
  \vspace{-1.0ex}
  \caption{Transition of control error of tool-tip position in tool-tip control experiment of PR2 with the connected duster.}
  \label{figure:pr2-connected-control-exp}
  \vspace{-3.0ex}
\end{figure}

\begin{figure}[t]
  \centering
  \includegraphics[width=1.0\columnwidth]{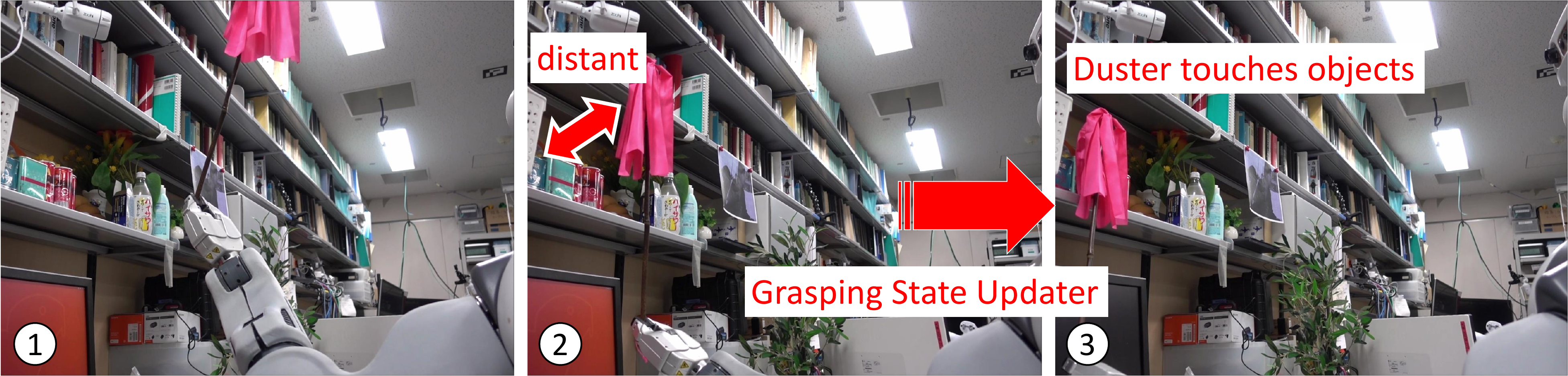}
  \vspace{-3.0ex}
  \caption{Duster-use experiment of PR2.}
  \label{figure:pr2-duster-exp}
  \vspace{-1.0ex}
\end{figure}

\subsection{PR2 Experiment} \label{subsec:pr2-exp}
\switchlanguage%
{%
  We perform experiments using the actual robot PR2.
  We perform the motion of \figref{figure:sim-hataki-exp} performed in \secref{subsec:simulation-exp} three times while changing the reference point of the tool-tip to obtain the data.
  The above procedure is repeated while changing the grasping state, and TBNPB is trained using about 1500 data points obtained, with 30 batches and 100 epochs.
  We fine-tuned the model obtained in \secref{subsec:simulation-exp} as described in \secref{subsec:training}.
  Since the grasping state is trained implicitly, we roughly create the states of holding the tool long or short and $\phi_{tool}=\{0, 30, 60\}$, and collect the data.
  The distribution of parametric bias obtained by fine-tuning is shown in \figref{figure:pr2-pb-exp}.
  In a similar but different form from \figref{figure:sim-pb-exp}, we can see that the parametric bias is neatly distributed along long or short $l_{tool}$ and $\phi_{tool}=\{0, 30, 60\}$.
  The results of the experiment on tool-tip control conducted in the same way as \figref{figure:sim-control-exp} are shown in \figref{figure:pr2-control-exp}.
  Initially, the control error is large, about 350 mm, because the grasping state is not known, but after the number of data exceeds $N^{online}_{thre}$ and the grasping state updater is executed, the control error suddenly decreases and drops to about 100 mm.
  After that, the control error did not change significantly even though we changed the grasping state by manually applying external force to the tool.
  The transition of parametric bias here is shown in ``trajectory'' of \figref{figure:pr2-pb-exp}, where (1) is the transition after the start of the updater and (2) is the transition after the change of the grasping state.
  We can see that parametric bias is automatically and correctly updated by detecting the change in the grasping state.

  Next, we perform an experiment using PR2 with the connected duster.
  As in the previous experiment, we collect data and train TBNPB, and the parametric bias obtained is shown in \figref{figure:pr2-connected-pb-exp}.
  The parametric bias is considered to have a form that varies more with $\phi_{tool}$ than with short / long $l_{tool}$, compared to \figref{figure:pr2-pb-exp}, because it bends greatly as the angle of the tool increases.
  The results of the same tool-tip control experiment as before are shown in \figref{figure:pr2-connected-control-exp}.
  The initial control error is about 890 mm, which is very large, but the grasping state gradually becomes known, and the error is reduced to about 160 mm.
  After that, the control error increases to about 450 mm when we change the grasping state by applying external force to the tool, but it decreases again to about 190 mm by the grasping state updater.
  The transition of the parametric bias here is shown in ``trajectory'' of \figref{figure:pr2-connected-pb-exp}, where (1) is the transition after the start of the updater and (2) is the transition after the change of the grasping state.
  For the flexible tool, we can see that the parametric bias is automatically and correctly updated by detecting the change in grasping state.

  Finally, the duster-use motion of PR2 with normal duster is shown in \figref{figure:pr2-duster-exp}.
  The duster-use motion is performed with a tool-tip position command such that the duster touches the objects on the shelves.
  At first, the duster does not touch the objects because the estimated grasping state is not correct, but after updating it, the duster correctly touches the objects and removes the dust.
}%
{%
  PR2の実機を使った実験を行う.
  \secref{subsec:simulation-exp}で行った\figref{figure:sim-hataki-exp}の動作を, 基準点を変えながら3回行いデータを取得する.
  把持状態を変化させながら上記を繰り返し, 得られた約1500のデータを使い, バッチ数を30, エポック数を100としてTBNPBを学習させる.
  このとき, \secref{subsec:training}で述べたように\secref{subsec:simulation-exp}で得られたモデルをfine-tuningする.
  また, 把持状態は暗黙的に学習されるため, 人間が目分量で, 道具を長く持った状態(long)と短く持った状態(short), $\phi_{tool}=\{0, 30, 60\}$の状態を作り, データを収集している.
  学習によって得られたparametric biasの分布を\figref{figure:pr2-pb-exp}示す.
  \figref{figure:sim-pb-exp}と似てはいるが異なる形で, long/short, $\phi_{tool}=\{0, 30, 60\}$で綺麗にpbが分布していることがわかる.
  また, \figref{figure:sim-control-exp}と同様の形で道具先端制御に関する実験を行った結果を\figref{figure:pr2-control-exp}に示す.
  初期は把持状態がわからないため, 制御誤差は約350 mmと大きいが, その後データ数が$N^{online}_{thre}$を超えてgrasping state updaterが実行されてから一気に制御誤差が減り, 制御誤差は約100 mmまで下がっていることがわかる.
  その後人間が手で道具に力を加えgrasping stateを変化させているが, 制御誤差に大きな変化は無かった.
  このときのpbの遷移は\figref{figure:pr2-pb-exp}の``trajectory''の通りであり, (1)がupdater開始後, (2)がgrasping stateの変化後の遷移である.
  grasping stateの変化を検知して自動でpbが更新されていることがわかる.

  次に, PR2 with connected dusterを使った実験を行う.
  先の実験と同様にデータを取得して学習し, その際に得られたparametric biasを\figref{figure:pr2-connected-pb-exp}に示す.
  道具が持つ角度によって大きく曲がるため, parametric biasも\figref{figure:pr2-pb-exp}に比べてshort/longに比べて$\phi_{tool}$によって大きく変化する形となっていると考えられる.
  先と同様の道具先端制御実験を行った結果を\figref{figure:pr2-connected-control-exp}に示す.
  初期の制御誤差は約1000 mmと非常に大きいが, 徐々にgrasping stateが分かっていき, 150 mm程度まで誤差が減っている.
  その後人間が道具に外力を加えてgrasping stateを変更すると500 mm程度まで制御誤差は増えるが, online updaterによりまた200 mm程度まで下がった.
  このときのpbの遷移は\figref{figure:pr2-connected-pb-exp}の``trajectory''の通りであり, (1)がupdater開始後, (2)がgrasping stateの変化後の遷移である.
  grasping stateの変化を検知して自動でpbが更新されていることがわかる.

  最後に, PR2 with normal dusterを使って行った実際のはたき動作を\figref{figure:pr2-duster-exp}に示す.
  壁のobjectsに接触するような道具先端位置指令を使ってはたき動作を行う.
  最初はgrasping stateが正しくないためはたきがobjectsに当たらないが, 更新を行うことで正しくobjectsにあたり, 埃を落とすことができた.
}%

\begin{figure}[t]
  \centering
  \includegraphics[width=0.8\columnwidth]{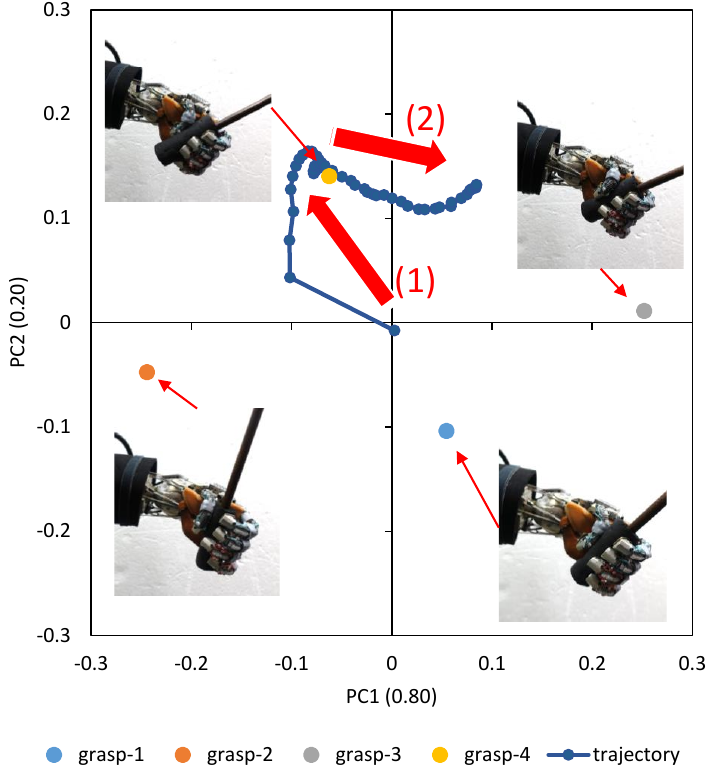}
  \vspace{-1.0ex}
  \caption{Parametric bias trained in MusashiLarm experiment and its trajectory in the tool-tip control experiment with grasping state updater.}
  \label{figure:musashilarm-pb-exp}
  \vspace{-3.0ex}
\end{figure}

\begin{figure}[t]
  \centering
  \includegraphics[width=1.0\columnwidth]{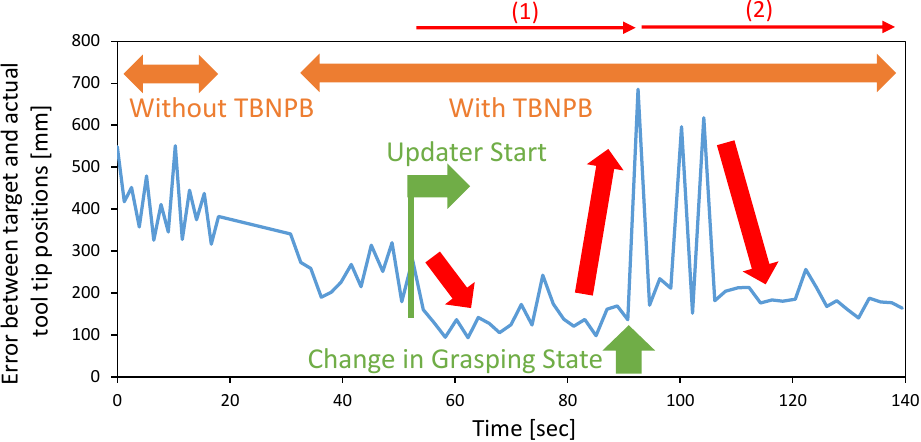}
  \vspace{-3.0ex}
  \caption{Transition of control error of tool-tip position in the tool-tip control experiment of MusashiLarm.}
  \label{figure:musashilarm-control-exp}
  \vspace{-3.0ex}
\end{figure}

\subsection{MusashiLarm Experiment} \label{subsec:musashilarm-exp}
\switchlanguage%
{%
  We perform experiments using the actual robot MusashiLarm.
  In this section, as in PR2, we first collect data using the geometric simulator of MusashiLarm and train TBNPB.
  Here, we fixed $l_{tool}=500$ [mm] and defined the angle $\psi_{tool}$ of the tool in the direction perpendicular to $\phi_{tool}$, and collected data with $\phi_{tool}=\{0, 30, 60\}$ [deg] and $\psi_{tool}=\{-30, 0, 30\}$ [deg].
  Note that $\bm{u}$ is 5-dimensional (2-dimensional wrist is not included) and $\bm{p}$ is 2-dimensional.
  After that, the data is collected by the actual robot as in \secref{subsec:pr2-exp} (in this case, joint angle commands are converted to muscle length by \cite{kawaharazuka2019longtime}), and the parametric bias after fine-tuning of TBNPB is shown in \figref{figure:musashilarm-pb-exp}.
  Unlike PR2, the grasping states are complex, so the human-created grasping states used for training are denoted as grasp-\{1, 2, 3, 4\}.
  The results of the tool-tip control experiment using the TBNPB as in PR2 are shown in \figref{figure:musashilarm-control-exp}.
  Using the geometric model with $\phi_{tool}=30$ and $\psi=0$ without TBNPB, the control error is about 410 mm, while using TBNPB, the control error is reduced to about 260 mm.
  In addition, by using grasping state updater, the control error is reduced to about 120 mm.
  After that, the grasping state is changed by applying an external force to the tool, and the control error is greatly increased, but it is reduced to about 180 mm again by the grasping state updater.
}%
{%
  MusashiLarmの実機を使った実験を行う.
  本節ではPR2と同様にまずMusashiLarmの幾何シミュレータによりデータを収集し学習させる.
  この際, $l_{tool}=500$ [mm]で固定し, $\phi_{tool}$と垂直な方向に関する道具の角度$\psi_{tool}$を定義し, $\phi_{tool}=\{0, 30, 60\}$ [deg], $\psi_{tool}=\{-30, 0, 30\}$ [deg]としてデータを収集した.
  なお, $\bm{u}$は5次元(手首2次元は入れていない)であり, $\bm{p}$は2次元とした.
  その後, PR2と同様に実機によりデータを取得し(この際, 関節角度指令は\cite{kawaharazuka2019longtime}により筋長に変換して行われる), TBNPBをfine-tuningした際のparametric biasを\figref{figure:musashilarm-pb-exp}に示す.
  PR2と違い把持状態が非常に複雑であるため, 学習に用いた人間が適当に作成した把持状態をgrasp-\{1, 2, 3, 4\}と表記している.
  このTBNPBを使ってPR2と同様に道具先端位置制御実験を行った結果を\figref{figure:musashilarm-control-exp}に示す.
  TBNPBを使わず$\phi_{tool}=30$, $\psi=0$とした幾何モデルを使った際は約410 mm程度の制御誤差があるのに対して, TBNPBを使うことで制御誤差が260 mm程度まで下がっている.
  また, さらにgrasping state updaterが加わることでgrasping stateが更新され, 制御誤差は120 mm程度まで下がった.
  その後道具に外力を加えることでgrasping stateが変化し, 制御誤差は大きく上昇するが, onlineによる更新により, また180 mm程度まで下がった.
}%

\section{Discussion} \label{sec:discussion}
\switchlanguage%
{%
  We discuss the results obtained from the experiments.
  From the simulator experiment of PR2, it is found that parametric bias self-organizes neatly according to the grasping state.
  They are self-organized in a way that is consistent with the fact that the longer the tool is, the larger the change of the tool-tip position due to the difference of the grasping angle is.
  We also found that the grasping state updater makes the current parametric bias transition to the correct value of the current grasping state.
  This allows the state estimation and the tool-tip position control to become more accurate than in the case without the grasping state updater.
  In addition, when the entire network is updated as in ordinary online learning, instead of updating only $\bm{p}$ as in this study, the tool-tip position can be controlled more accurately around the learned data, but over-fitting occurs and the control error becomes larger for the untrained data.
  Since our grasping state updater updates only the grasping state, not the entire network, it is possible to reduce the control error for untrained data.
  The same tendency is observed in the actual robot of PR2, and it is possible to update the grasping state by the online updater and to reduce the control error with it.
  Similar results are obtained in the experiment using the connected duster, which is a flexible tool, and it is found that the method can be applied not only to rigid tools but also to deformable tools.
  Finally, in the experiment using MusashiLarm, we dealt with a system in which the grasping state is more ambiguous and the body is flexible.
  It is found that the control error is very large without TBNPB, but it is reduced by using TBNPB, and it is further reduced by updating the ambiguous grasping state.
  In other words, TBNPB can be applied to flexible hands where the grasping state cannot be defined and to flexible robots where the joint angle cannot be realized precisely.
  In addition, depending on how the data are collected, irreproducible initialization and deterioration over time, which are specific to flexible bodies, can be included in the parametric bias as in \cite{kawaharazuka2020dynamics}.
  Therefore, it is found that the estimation and control of the tool-tip, and update of the grasping state are possible for rigid axis-driven robots, flexible tendon-driven robots, various robotic hands, rigid tools, and deformable tools.

  The main limitations of this study are (1) data collection, (2) range of tool types, and (3) control error.
  Regarding (1), since we have to obtain data for each tool, the current method does not scale up to many tools.
  On the other hand, it should be possible to embed the tool types into the parametric bias as well as the grasping state.
  In this case, we need to obtain a large amount of data (i.e., various tool types and various grasping states), but we can infer new tools and grasping states that correspond to the internally dividing points of the training data.
  In addition, we think that obtaining a large amount of prior tool data by using simulation is one direction.
  Regarding (2), this study can currently handle rigid and elastic tools, but it is difficult to use tools with melting or breaking properties.
  In this study, the mapping from $\bm{u}$ to $\bm{x}_{tool}$ is embedded in the weights $W$, and the rest is embedded in $\bm{p}$.
  Therefore, tools such as rigid and elastic bodies, for which $\bm{x}_{tool}$ can be calculated from $\bm{u}$ using the effect of gravity or the structure of the tool, can be treated in the same way as in this study.
  On the other hand, for tools with melting or breaking properties, where $\bm{x}_{tool}$ is not uniquely determined from $\bm{u}$ due to transformations, those transformations will be embedded in $\bm{p}$.
  In this case, the changes in the tool structure and the grasping state will be embedded in the same $\bm{p}$, which is likely to make this study less useful.
  Regarding (3), since the tool-tip shape of the duster is amorphous and the observation error is large, the error of the experimental results is relatively large.
  Although this was not a problem because the duster does not require such precise operation, the control error becomes a big problem when handling tools such as drills and saws.
  The inference accuracy of our network mainly depends on the recognition accuracy of the tool-tip and the motion range of the tool-tip in the whole operation.
  Since the smaller the whole motion is, the relatively higher the inference accuracy for the tool-tip position will be, we think that drills and hammers can be handled by making the whole motion small.
  Since our method learns the relationship between the body and the tool itself, once the grasping state is correctly updated, the robot can look away from the tool, which is the difference from visual feedback, in which the robot must continue to look at the tool.
  However, if we want to perform more precise motions, we should consider using TBNPB for tool-tip control to some extent, and then using it together with visual feedback.

  The application of this research is not limited to the tool-tip control.
  For a group of sensors and actuators that have some static relationships, it is possible to learn and use these relationships while embedding implicit and difficult-to-observe information into parametric bias by using this study.
  This is the first method that brings parametric bias, which has been used in imitation learning, to static relationships.
  Moreover, since it uses a neural network, it is very easy to integrate not only the relationship between two sensors but also other sensors.
  In the future, we would like to learn the relationship between various sensors, including contact sensors and torque sensors.
  We would also like to try new tasks such as controlling the water coming out of a hose.
}%
{%
  実験から得られた結果について議論する.
  PR2のシミュレータ実験からは, 綺麗に特性ごとにpbが自己組織化されることがわかった.
  それらは道具が長いほど角度の違いによる道具先端位置の変化が大きいというような事実と一致した形で自己組織化される.
  また, grasping state updaterにより, pbが現在のgrasping stateのparametric bias周辺に遷移していくことがわかった.
  これにより, 状態推定・道具先端位置制御がgrasping state updaterがない場合に比べて正確になっていくことがわかった.
  また, 本研究のように$\bm{p}$のみを更新するのではなく, 通常のオンライン学習のようにネットワーク全体を更新した場合は, 学習したデータ付近についてはより正確に道具先端位置制御が可能となるものの, 過学習が起き, 学習していないデータについては, 制御誤差が大きくなってしまうことがわかった.
  本研究のgrasping state updaterはネットワーク全体ではなく把持状態のみを更新するため, 学習していないデータについても制御誤差を減少させることが可能である.
  PR2の実機においても同様の傾向が見られ, updaterによるgrasping stateの更新, それによる制御誤差の減少が可能であった.
  柔軟な道具であるconnected dusterを用いた実験でも同様の結果が得られ, 剛体の道具だけでなく, 変形する道具にも適用可能であることがわかった.
  最後にMusashiLarmを使った実験では, grasping stateがより曖昧かつ, 身体が柔軟な系を扱った.
  身体が柔軟であるため指令した関節角度を正確に実現できるわけではないため, TBNPBを使わない場合は非常に制御誤差が大きいが, TBNPBを使うことで制御誤差が減り, さらにその曖昧なgrasping stateを更新することでより制御誤差が減ることがわかった.
  つまり, grasping stateが定義できないような系や, 関節角度が正確に実現できないような柔軟系にも適用することができる.
  また, データの取り方次第では柔軟身体に特有なirreproducible initializationや経年劣化についても, parametric biasに含めることが可能であると考えられる\cite{kawaharazuka2020dynamics}
  よって, 剛な軸駆動系, 柔軟な腱駆動系, 剛な道具, 変形する道具, 様々なハンドに対して, 道具先端位置の推定・制御・把持状態の更新が可能であることがわかった.

  本研究の主なlimitationは(1)データ取得, (2)道具の種類, (3)制御誤差である.
  (1)について, 現状道具ごとにそれぞれデータを取得しないといけないため, 道具の種類についてスケーリングしない形になっている.
  一方, 道具の種類についても把持状態と同様にparametric biasに埋め込むことは可能なはずである.
  この場合, 道具の種類 x 様々な把持状態という大量のデータを取得する必要があるが, それらの内分点にあたるような道具や把持状態は推論可能になると考える.
  また, シミュレーションの活用による大量の事前データの取得も一つの方向性だと思う.
  (2)について, 現状剛体や弾性体の道具は使うことができるが, meltingやbreakingを伴う道具の利用は難しい.
  剛体や弾性体のように, 重力や道具の構造によって$\bm{u}$から$\bm{x}_{tool}$が推論可能な道具は本研究と同様に扱うことができる.
  一方, meltingやbreakingのように, 変形によって, $\bm{u}$から$\bm{x}_{tool}$が一意に定まらなくなってしまう道具については, それらの変化が$\bm{p}$に埋め込まれることになる.
  この場合, 道具の変化と把持状態が同じ$\bm{p}$に埋め込まれてしまうため, 使い勝手の良くないものになってしまう可能性が高く, 今後検証していく必要がある.
  (3)について, dusterは道具先端位置が不定形であり観測誤差が大きいため実験結果の誤差もそれなりに大きい.
  また, そこまで正確な操作を必要とする道具ではないため問題は無かったが, ドリルやノコギリ等を扱う際は制御誤差が大きな問題となる.
  本研究のネットワークの推論精度は主に, 道具先端の認識精度と, 動作全体における道具先端の動きの大きさに依存する.
  動作全体が小さければ小さいほど, 相対的に道具先端位置に対する推論精度が上がるため, 同様にドリル等も扱うことができると考えている.
  本手法は身体と道具の間の関係自体を学習するため, 一度把持状態が更新されれば, 後はよそ見をして別の方向を見たりすることも可能であり, これがvisual feedbackとの違いである.
  しかし, より精密な動作を行う場合, TBNPBである程度の道具制御をし, その後visual feedbackと併用して用いることも考える必要がある.

  本研究の適用先は道具制御だけには留まらない.
  何らかの静的な関係性を持ったセンサ・アクチュエータ群に対して, 暗黙的で観測の難しい情報をparametric biasに押し込めながら, その関係性を学習して利用するものである.
  これまでimitation learning等において使われていたparametric biasを初めて静的な関係に持ち込んだ手法である.
  また, Neural Networkを使用しているため2つのセンサ関係性をだけでなく, 他のさらなるセンサとの融合も非常に簡単である.
  今後はさらに接触センサや関節トルク等の次元を入れた様々なセンサ間の関係を学習していきたい.
  また, ホースの水のかかる位置制御等, 新しいタスクに挑戦していきたい.

}%

\section{Conclusion} \label{sec:conclusion}
\switchlanguage%
{%
  In this study, we constructed a network to estimate the tool-tip position from the body control command, and developed a method to control the tool-tip position using backpropagation.
  By including parametric bias, implicit variables related to the grasping state are embedded in the network, and the network can adapt to the online changes in the grasping state through online learning.
  In addition, by using a neural network instead of a linear transformation, we have confirmed that the system can handle deformable tools and bodies with different structures such as axis-driven and tendon-driven robots.
  In the future, we would like to discuss the extension to the dynamic tool-use and the integration with redundant sensors such as contact and torque sensors.
}%
{%
  本研究では, 身体の制御入力から道具先端位置を推定するネットワークを構成し, 誤差逆伝播を用いて道具先端位置を制御する手法を開発した.
  Parametric Biasを含めることで, その中に把持状態に関する暗黙的な変数を埋め込み, これをオンライン学習することで, 逐次的な把持状態変化に適応することができる.
  また, アフィン変換ではなくニューラルネットワークにより表現することで弾性変形するような道具・軸駆動や筋骨格などの異なる身体も同様に扱うことができることを確認した.
  今後, 動的な道具利用への拡張, 接触センサ等の冗長なセンサとの統合的利用についても議論していきたい.
}%

{
  \bibliographystyle{IEEEtran}
  \bibliography{main}
}

\end{document}